\pdfoutput=1

\documentclass[11pt]{article}

\usepackage{hyperref}
\usepackage{url}
\usepackage{array,graphicx}
\usepackage{booktabs}
\usepackage{multirow}
\usepackage{wrapfig}
\usepackage{float}
\usepackage{tabularx}
\usepackage{subcaption}
\usepackage{amsmath}
\DeclareMathOperator*{\argmax}{arg\,max}

\usepackage{acl2023}

\usepackage{times}
\usepackage{latexsym}

\usepackage[T1]{fontenc}

\usepackage[utf8]{inputenc}

\usepackage{microtype}

\title{SCOTT:  Self-Consistent Chain-of-Thought Distillation}

\author{Peifeng Wang$^{1}$\thanks{~~This work was done when Peifeng Wang was an intern at Amazon.  Zheng Li and Xiang Ren are corresponding authors.}, Zhengyang Wang$^{2}$, Zheng Li$^{2}$, Yifan Gao$^{2}$, Bing Yin$^{2}$, Xiang Ren$^{1}$\\
$^{1}$Department of Computer Science, University of Southern California, $^{2}$Amazon.com Inc\\
\texttt{\{peifengw,xiangren\}@usc.edu},\\
\texttt{\{zhengywa,amzzhe,yifangao,alexbyin\}@amazon.com}}

\begin{document}
\maketitle
\begin{abstract}
Large language models (LMs) beyond a certain scale, demonstrate the emergent capability of generating free-text rationales for their predictions via chain-of-thought (CoT) prompting.
While CoT can yield dramatically improved performance, such gains are only observed for sufficiently large LMs. 
Even more concerning, there is little guarantee that the generated rationales are consistent with LM's predictions or faithfully justify the decisions. 
In this work, we propose SCOTT, a faithful knowledge distillation method to learn a small, self-consistent CoT model from a teacher model that is orders of magnitude larger. 
To form better supervision, we elicit rationales supporting the gold answers from a large LM (teacher) by contrastive decoding, which encourages the teacher to generate tokens that become more plausible only when the answer is considered. 
To ensure faithful distillation, we use the teacher-generated rationales to learn a student LM with a counterfactual reasoning objective, which prevents the student from ignoring the rationales to make inconsistent predictions. 
Experiments show that, while yielding comparable end-task performance, our method can generate CoT rationales that are more faithful than baselines do. 
Further analysis suggests that such a model respects the rationales more when making decisions; thus, we can improve its performance more by refining its rationales.
\footnote{Code can be found at~\url{https://github.com/wangpf3/consistent-CoT-distillation}.}

\end{abstract}

\section{Introduction}


Large language models (LMs) elicit strong reasoning capabilities through chain-of-thought (CoT) prompting~\cite{wei2022chain}, which asks LMs to generate free-text rationale for explaining their multi-step reasoning.
However, CoT prompting does not guarantee that the rationale is consistent with the prediction, rendering the rationale useless for justifying the model’s behavior. In this work, we present \textbf{S}elf-Consistent \textbf{C}hain-\textbf{O}f-\textbf{T}hought Dis\textbf{T}illation (\textbf{SCOTT}), a knowledge distillation (KD) method for eliciting faithful CoT reasoning, 
where a small student model learns from a large teacher model to generate CoT rationales that are consistent to its own predictions.

\begin{figure}[t]
    \centering
    \includegraphics[width=0.5\textwidth]{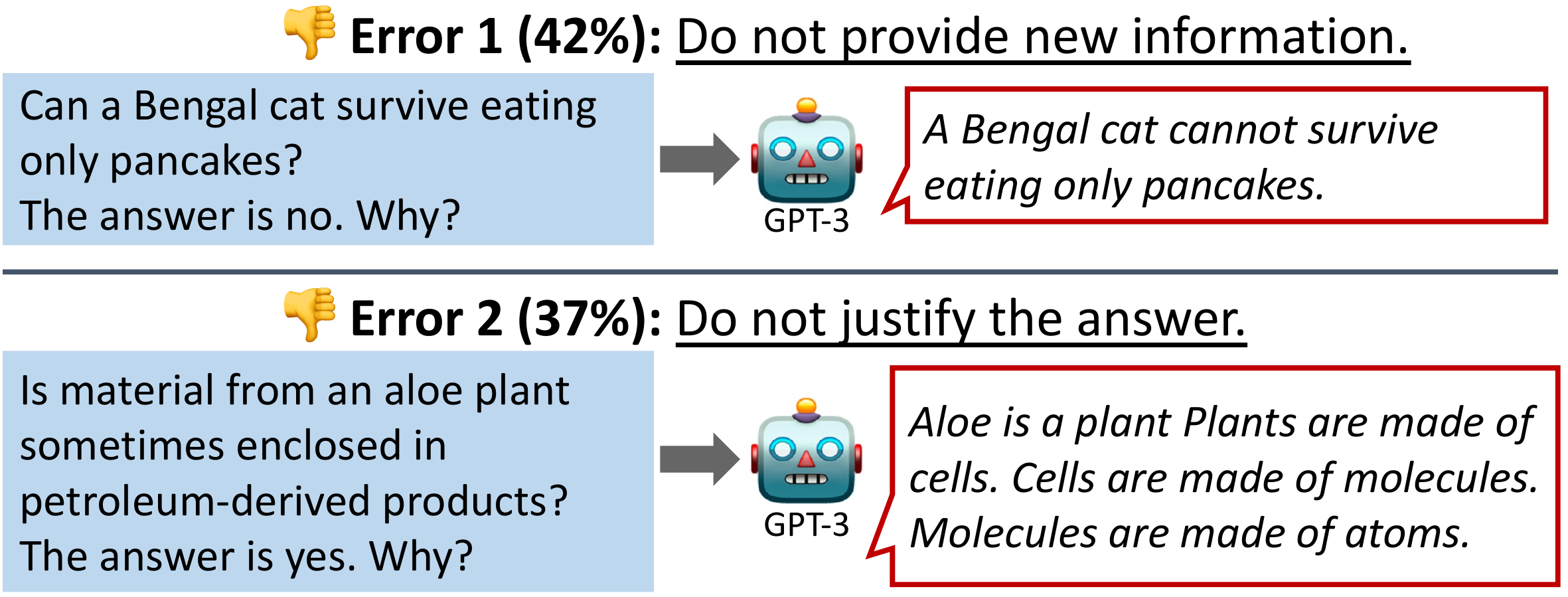}
    \caption{Vacuous rationales generated by a prompted LM (GPT-3) for StrategyQA. In both types of error cases, LM fails to give rationales consistent with the answers due to hallucination.}
    \label{fig:vacuous_rationale}
\end{figure}

\begin{figure*}[ht]
    \centering
    \includegraphics[width=0.9\textwidth]{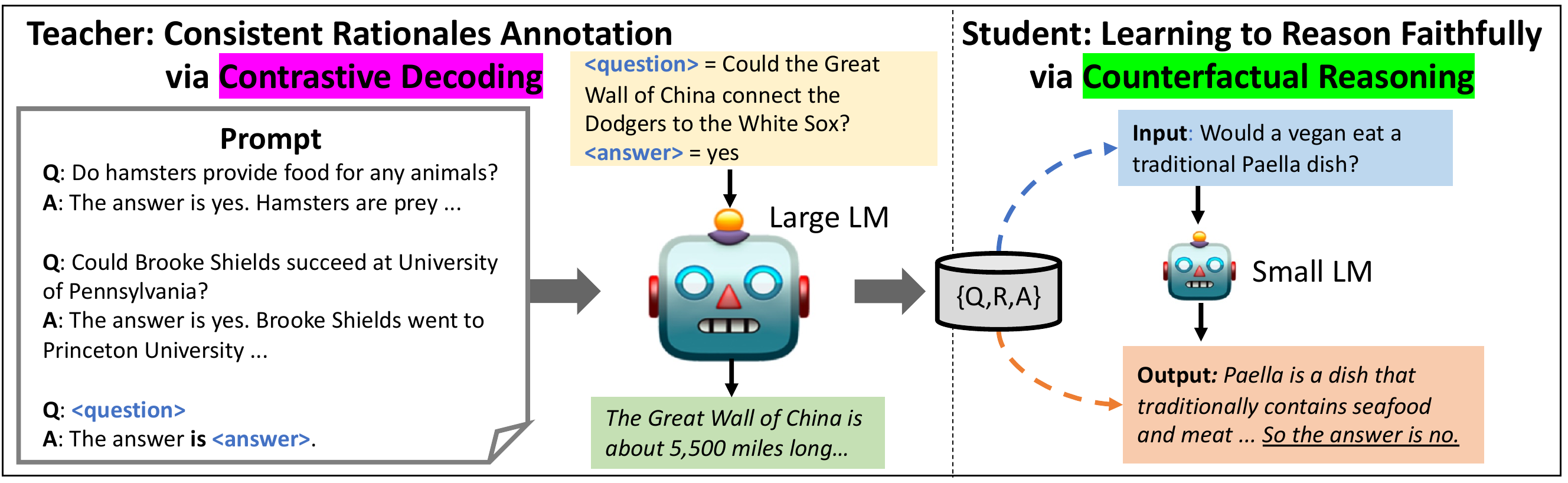}
    \caption{\textbf{Overview of our knowledge distillation framework for faithful reasoning.} (a) Teacher: A large LM prompted to generate a consistent rationale given a question and the gold answer in the training set via contrastive decoding. (b) Student: A small LM fine-tuned to generate a rationale and then answer via counterfactual reasoning.
    }
    \label{fig:pipeline}
\end{figure*}
Existing works~\cite{shridhar2022distilling,li2022explanations} propose learning to reason from large LMs mainly for computation efficiency or task performance. They prompt a large LM (the teacher) to generate rationales for a downstream dataset, which is then used to train a small LM (the student). 
However, these works neglect the following two issues which could undermine the faithfulness of the rationales.
First, LMs are prone to hallucination, meaning they often generate text that is not grounded by the input~\cite{maynez-etal-2020-faithfulness,ji2022survey}. Therefore,  the teacher may not generate on-topic rationales, which fully support the answer. In our pioneer study (Figure~\ref{fig:vacuous_rationale}) over 100 random rationales generated by GPT-3, we found $42\%$ of them not providing new information that is not stated in the task input and $37\%$ of them not justifying the answer\footnote{\citeauthor{wiegreffe2021reframing} obtains a similar observation on the rationales generated by GPT-3 for the CommonsenseQA dataset.}. This inconsistency between the rationale and answer would then be inherited by the student. Second, the student may treat rationale generation and answer prediction as two independent processes. This is due to the spurious correlations between the question and answer, which is exploited as a reasoning shortcut by the student~\cite{branco2021shortcutted}. The two issues together would lead to an unfaithful student which learns to generate vacuous rationales and may make  predictions inconsistent with the rationales.

To address these issues, we propose to enhance the vanilla KD process from two ends respectively. To elicit more on-topic rationales from the teacher, we propose to leverage contrastive decoding which aims to ground each rationale to the answer (\S~\ref{constrative_decoding}). This technique encourages the teacher to generate tokens that are more plausible only when the answer is considered instead of the ones that are fairly plausible even without the answer during decoding. To train a faithful student, we ask the student to conduct counterfactual reasoning, i.e., predicting accordingly when the rationales are leading to different answers (\S~\ref{sec:counterfactual_reasoning}). We obtain the training data by asking the teacher to generate a rationale for a sampled incorrect answer. The reasoning shortcut between the question and the gold answer is thus removed since now the student needs to give a different answer for the same question, according to the rationales provided during training. 

We conduct experiments on several open-domain question answering tasks that require knowledge-intensive reasoning.
Experiments show that: (1) Contrastive decoding can lead to a more consistent teacher which generates rationales that are more supportive of the gold answers. (2) Trained on the more consistent rationale-answer pairs, the student learns to better associate the answer prediction with the rationale generation. (3) With counterfactual reasoning as an auxiliary training objective, the student learns not to take the reasoning shortcut and instead respect the rationale more. (4) Despite being more faithful, our model performs comparably to the baselines. (5) Ablation study shows that although performing better, larger student models are more prone to being inconsistent.  Our method robustly remedies the inconsistency regardless of the size of the student model. (6) With a more faithful student, we can better improve its performance by correcting its rationale, demonstrating the utility of our method in model refinement.

\section{Chain-of-Thought Distillation}

Our goal is to 1) elicit consistent rationales, i.e., those well justifying the gold answers, from a large LM as supervision, and then 2) train a self-consistent student model to reason faithfully, i.e., answer accordingly to its generated rationale. 
We consider the task of language-based reasoning where the required knowledge is not provided in the task input. Specifically, we focus on open-domain question answering (QA) which is the most general setting adopted by prior works: given a question $q$, a QA system is asked to predict the gold answer $a^*$. For interpretability, we also require the model to provide a free-text rationale $r$, which justifies its prediction.
Below we describe the overview of a vanilla KD framework as illustrated in Figure~\ref{fig:pipeline}. We then discuss the limitations and propose our method in \S~\ref{sec:method}.

\subsection{Generating Rationale Annotation}
Instead of asking humans to annotate a rationale for each question-answer tuple $\{q, a^*\}$, we obtain the rationale from a teacher model automatically using in-context learning. The idea is to prompt a frozen LM as the teacher with only a few annotated examples as demonstration before a new instance is provided. Each example consists of a question $q$ randomly sampled from the training set, 
the gold answer $a^*$ and a human-annotated rationale $r$ which justifies why $a^*$ is correct. The prompt $p$ is structured in the format as shown in Figure~\ref{fig:pipeline} (the Prompt in the left part). To obtain the rationale for a new question $q$, one basic strategy could be greedy decoding, which selects the most plausible token at each step:
\begin{equation}\label{eq:greedy_decoding}
    t_i^{*}=\argmax \log P(t_i|p,q,a^*,t_{<i}).
\end{equation}

\subsection{Training a Student Model}
Now with the annotated training data $\{q,r,a^*\}$, we can train a smaller model as the student. There are many ways to implement a QA model that can make a prediction as well as generate a rationale. In this work, we focus on the self-rationalization paradigm, where the student firstly generates a rationale and then predicts the answer conditioning on the generated rationale. This is in contrast to related works which conduct post-rationalization, i.e., generating the rationale after the answer is predicted, or multi-task learning, which treats rationale generation as an auxiliary task besides answer prediction. The reason is that the generation of the rationale for the latter two paradigms does not affect the decision making by design, and therefore the faithfulness of the rationale is not guaranteed in the first place.  

Given a question $q$,
the student model is trained to output a sequence of rationale tokens concatenated with the answer tokens as shown in Figure~\ref{fig:pipeline} (the output in the right part). One straightforward implementation is simply fine-tuning a text-to-text LM over the silver training data generated by the teacher using standard language modeling loss:
\begin{equation}\label{eq:factual_training}
    \mathcal{L}_{factual}=-\sum_i\log P(t_i|q,t_{<i}),
\end{equation}
which we refer as factual reasoning loss.
\section{Distilling a Self-Consistent Student}
\label{sec:method}

There are two vital issues with the vanilla KD process described in the previous section. Firstly, neural LMs are known to suffer from the issue of hallucination, meaning they often generate text that is not grounded by the input~\cite{maynez-etal-2020-faithfulness,ji2022survey}. This would lead to the generated rationale not supporting the given answer. The inconsistency between the rationale and the answer would then be inherited by the student, which is misled to think that the answer prediction is independent of the rationale generation. Secondly, the student model would learn to predict the answer by taking a reasoning shortcut~\cite{branco2021shortcutted}, without taking into account the generated rationale (even though the answer prediction is conditioned on the rationale). This is due to the spurious correlations between the question and the answer which are found in various implicit reasoning task datasets~\cite{gururangan2018annotation,zellers2019hellaswag,blodgett2020language}.

The two issues mentioned above would result in an untrustworthy student whose generated rationales do not consistently justify its answers. To mitigate this, we propose two corresponding techniques as detailed below.

\subsection{A Consistent Teacher: Contrastive Decoding}\label{constrative_decoding}
\begin{figure}[t]
    \centering
    \includegraphics[width=0.5\textwidth]{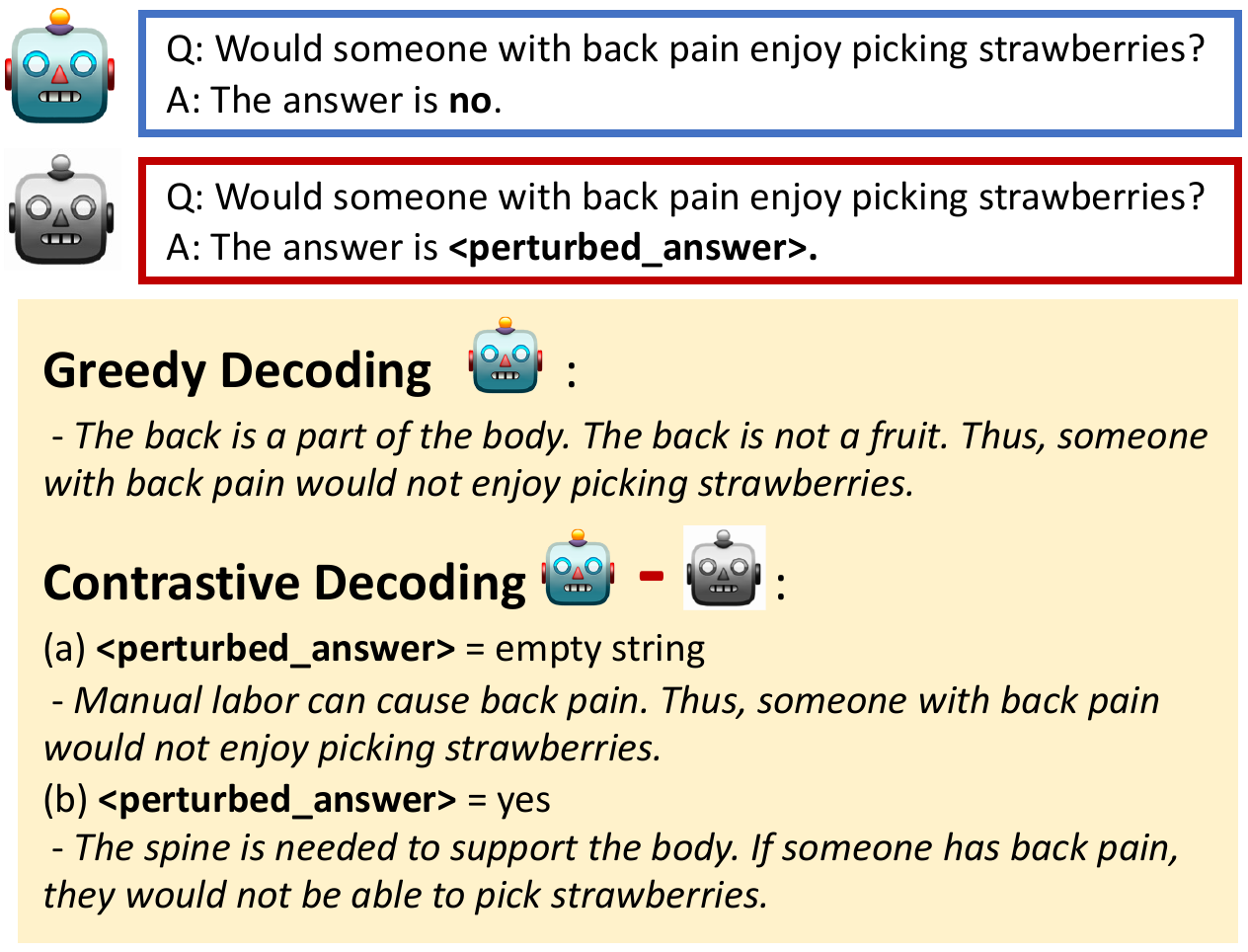}
    \caption{\textbf{Contrastive decoding} for obtaining rationales that are more grounded by the gold answers, by preferring tokens that are more plausible only when the answer is considered. 
    }
    \label{fig:contrastive_decoding}
\end{figure}
To encourage the teacher to generate a more on-topic rationale that supports the answer, our proposed method extends a prior technique called contrastive decoding for open-ended text generation~\cite{li2022contrastive}. The core idea is to search rationale tokens 
that are more plausible only when the answer is considered instead of the ones that are fairly plausible even without the answer during decoding.
To implement this idea, we firstly model the hallucinating behavior by providing a perturbed answer $a^{'}$ to the same teacher and then obtain the plausibility growth of any token $t_i$ given the answer $a^*$ as 
\begin{equation}\label{eq:plausibility_growth}
    G(t_i|a^*)=\log\frac{P(t_i|p,q,a^*,t_{<i})}{P(t_i|p,q,a^{'},t_{<i})}.
\end{equation}
We investigate two ways of perturbing the answer: setting $a^{'}$ as an empty string or an incorrect answer other than $a^*$\footnote{For yes/no or true/false questions, we obtain the incorrect answer by flipping the gold answer. For multi-choice questions, we randomly pick one incorrect answer.}. The first way (with an empty string) punishes tokens that are generally plausible when the gold answer $a^*$ is not considered by a hallucinated LM. The second way (with an incorrect answer) takes a step further by encouraging the teacher to generate a rationale that is more distinctive between gold and wrong answers. Figure~\ref{fig:contrastive_decoding} shows the generations for an example question from greedy decoding and contrastive decoding.

To strike a balance between language fluency and the grounding with $a^*$, we incorporate the plausibility growth into Eq.~\ref{eq:greedy_decoding} by aggregation as our final contrastive decoding strategy:
\begin{equation}\label{eq:contrastive_decoding}
    t_i^{*}=\argmax\log P(t_i|p,q,a^*,t_{<i})+G(t_i|a^*)
\end{equation}

\subsection{A Faithful Student: Counterfactual Reasoning}\label{sec:counterfactual_reasoning}
\begin{figure}[t]
    \centering
    \includegraphics[width=0.49\textwidth]{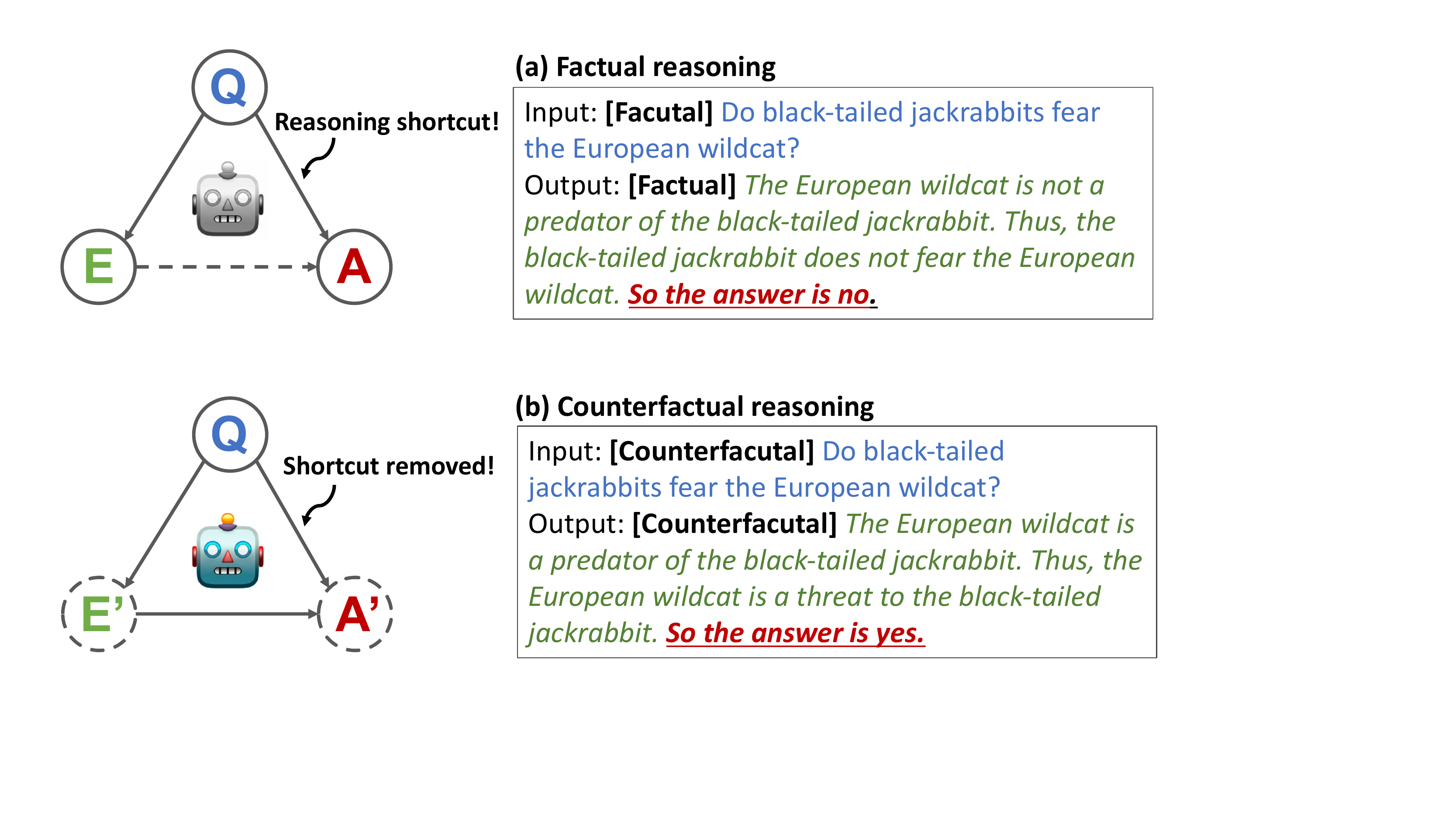}
    \caption{\textbf{Counterfactual reasoning} for teaching the student to reason faithfully, i.e., answer differently according to the rationale. 
    }
    \label{fig:counterfactual_training}
\end{figure}

To encourage the student to reason faithfully towards its generated rationale, we train the student to conduct counterfactual reasoning~\cite{roese1997counterfactual}, i.e., answer accordingly when the rationale is leading to a different answer. This would help remove the reasoning shortcut between a question and the gold answer (Figure~\ref{fig:counterfactual_training}) since now the student is asked to answer differently for the same question. To implement this idea, we firstly replace the gold answer fed to the teacher in Eq.~\ref{eq:contrastive_decoding} with a wrong answer $a^{'}$ randomly (with the same sampling strategy as in \S~\ref{constrative_decoding}) as if $a^{'}$ is correct. We thus obtain a counterfactual rationale $r^{'}$ that leads to the wrong answer $a^{'}$. We then train the model to generate $a^{'}$ when $r^{'}$ is directly fed to the decoder as teacher-forcing (the language modeling loss is only applied to the answer tokens $t_i\in a^{'}$):
\begin{equation}\label{eq:counterfactual_training}
    \mathcal{L}_{counterfactual}=-\sum_i\log P(t_i|q,r^{'},t_{<i}).
\end{equation}

To avoid confusing the student about the task, we indicate the training objective Eq.~\ref{eq:factual_training} (or Eq.~\ref{eq:counterfactual_training}) to the student by appending the keyword \texttt{[Factual]} (or \texttt{[Counterfactual]}) at the beginning of both the input sequence to the encoder and the output sequence to the decoder (see Figure~\ref{fig:counterfactual_training} for an example input and output). The overall training loss is the sum of Eq.~\ref{eq:factual_training} and Eq.~\ref{eq:counterfactual_training}.
\section{Experiments}
%
We aim to answer the following research questions in our experiments: (1) Can our contrastive decoding strategy lead to a more consistent teacher?
(2) Can a more consistent teacher and the counterfactual reasoning objective lead to a student that reasons more faithfully?
(3) Can we have more control over a self-consistent student's predictions by modifying its generated rationales?

\begin{figure*}[t]
    \centering
    \includegraphics[width=1.0\textwidth]{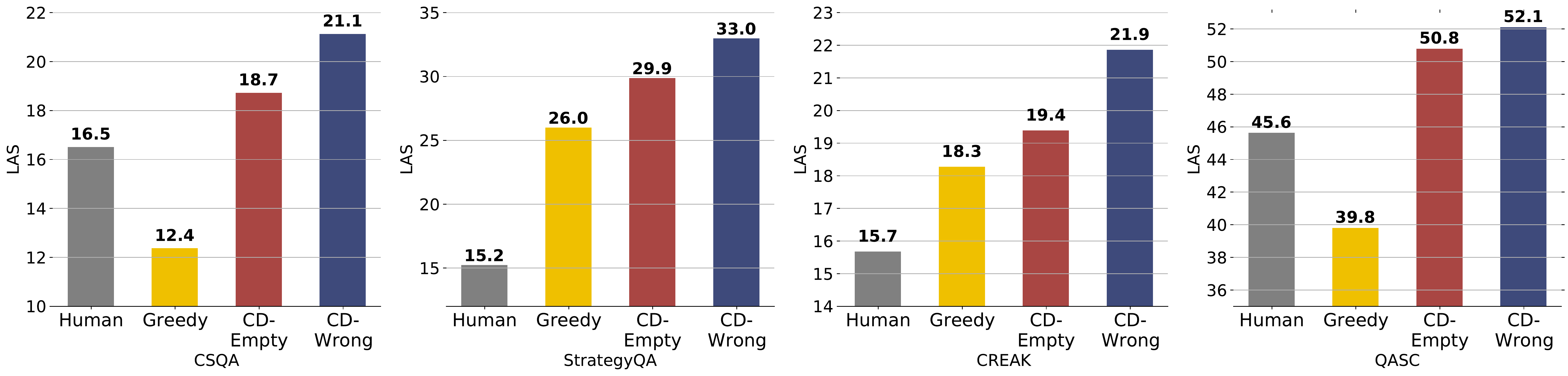}
    \caption{Simulatability (LAS) of the rationales generated from different teacher models as a measurement the consistency between the rationales and the gold answers. \{\textit{Greedy, CD-Empty, CD-Wrong}\} refer respectively to using greedy decoding, contrastive decoding with empty/wrong answer to obtain rationale tokens from the teacher.
    }
    \label{fig:teacher_las}
\end{figure*}
\subsection{Datasets}
We experiment with several language-based reasoning tasks that are knowledge-intensive:
(1) CSQA~\citep{talmor2018commonsenseqa} is a five-choice QA dataset that tests general commonsense about the daily concepts.
(2) StrategyQA~\citep{geva2021did} is a binary 
(yes/no) QA dataset where the required reasoning steps are implicit in the question. (3) CREAK~\citep{onoe2021creak} is a fact-checking (true/false) dataset which tests commonsense reasoning about entity knowledge.
(4) QASC~\citep{khot2020qasc} is an eight-choice QA dataset which requires both knowledge facts retrieval and the common sense for composing the facts. 
Since the test labels for these datasets are not publicly available, we treat the official development set as our test set, while randomly splitting the official training set into a new training set and development set.

\subsection{Evaluation Metrics}
(1) To evaluate the consistency between the rationales generated by the teacher and the gold answers provided as input, we use the LAS metric~\cite{hase2020leakage}, whose core idea is to measure how well the rationales assist a simulator to predict the \textit{gold answers $a^{*}$}, computed as the difference between the task performance when the rationale is provided as input vs. when it is not, namely $Acc(qr\rightarrow a^*)-Acc(q\rightarrow a^*)$.
(2) To evaluate the faithfulness of the rationales generated by the student, we use LAS to measure how well the rationales help a simulator to predict \textit{a student's predictions $a^{'}$}, namely $Acc(qr\rightarrow a^{'})-Acc(q\rightarrow a^{'})$. 
We implement each simulators with a fine-tuned T5-large model~\cite{2020t5} respectively.
(3) To evaluate how well the student preserves its task performance on the downstream datasets, we use accuracy as the metric.  

\subsection{Implementation Details}
We use GPT-neox~\cite{gpt-neox-20b}, a LM with 20B parameters as the teacher since the model checkpoint is publicly available, which allows us to host it offline and have access to token-wise probabilities as required in our contrastive decoding. We then implement two teacher variants by using an empty string or a wrong answer
as the perturbed answer $a^{'}$ in Eq.~\ref{eq:contrastive_decoding} respectively. The obtained rationales are then used to fine-tune two T5-3b LMs as the students 
respectively.
For both variants, we train the student using the sum of factual training loss Eq.~\ref{eq:factual_training} and counterfactual training loss Eq.~\ref{eq:counterfactual_training}. 

\subsection{Baselines}

\textbf{Chain-of-Thought (CoT)} 
Since we elicit the rationales from GPT-neox (with 20b parameters)~\cite{gpt-neox-20b} to train the student, we prompt the same model (GPT-neox) to firstly explain and then predict using CoT prompting~\cite{wei2022chain}. 

\smallskip
\noindent
\textbf{Learn from Human} To demonstrate the advantage of our automatic way of generating rationale annotations, we implement this baseline as a fine-tuned T5-3b LM over human-annotated rationales, which are expensive to obtain and could be noisy. 

\smallskip
\noindent
\textbf{Learn from Greedy Decoding} We implement this baseline as a fine-tuned T5-3b LM over the rationales  obtained by greedy decoding using the same LM as our main method. We also implement another variant by adding the counterfactual reasoning loss when fine-tuning the student, where the rationales for the wrong answers are obtained by greedy decoding.

We also implement two baselines of our method 
by training the student with the rationales obtained by contrastive decoding with empty/wrong answers based on factual reasoning only. We run all the experiments  for 5 times using a fixed set of random seeds and report the average results.

\subsection{Main Results}

\begin{figure*}[ht]
    \centering
\begin{subfigure}{.45\textwidth}
  \centering
  \includegraphics[width=\linewidth]{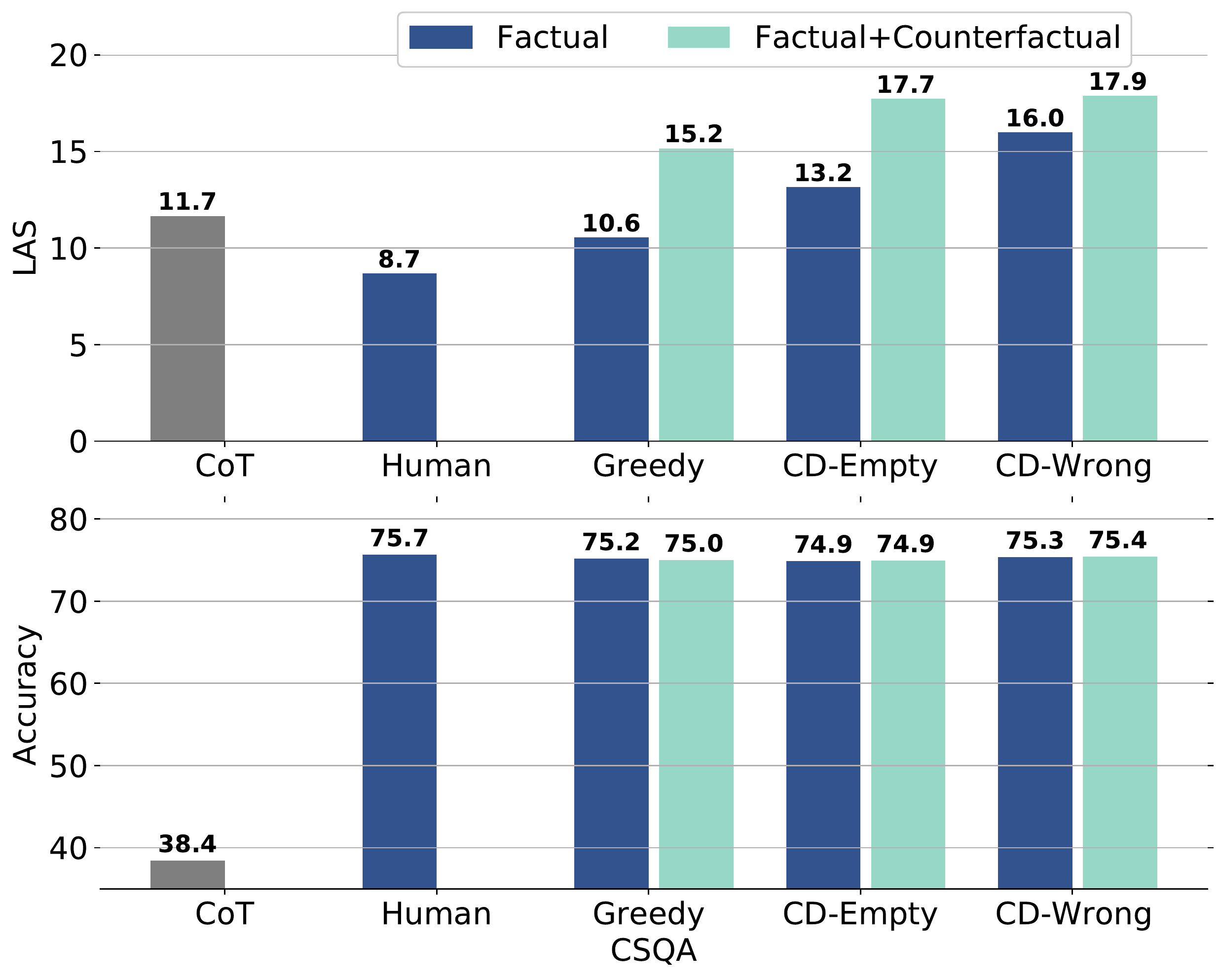}  
  \label{fig:acc_las_csqa}
\end{subfigure}
\begin{subfigure}{.45\textwidth}
  \centering
  \includegraphics[width=\linewidth]{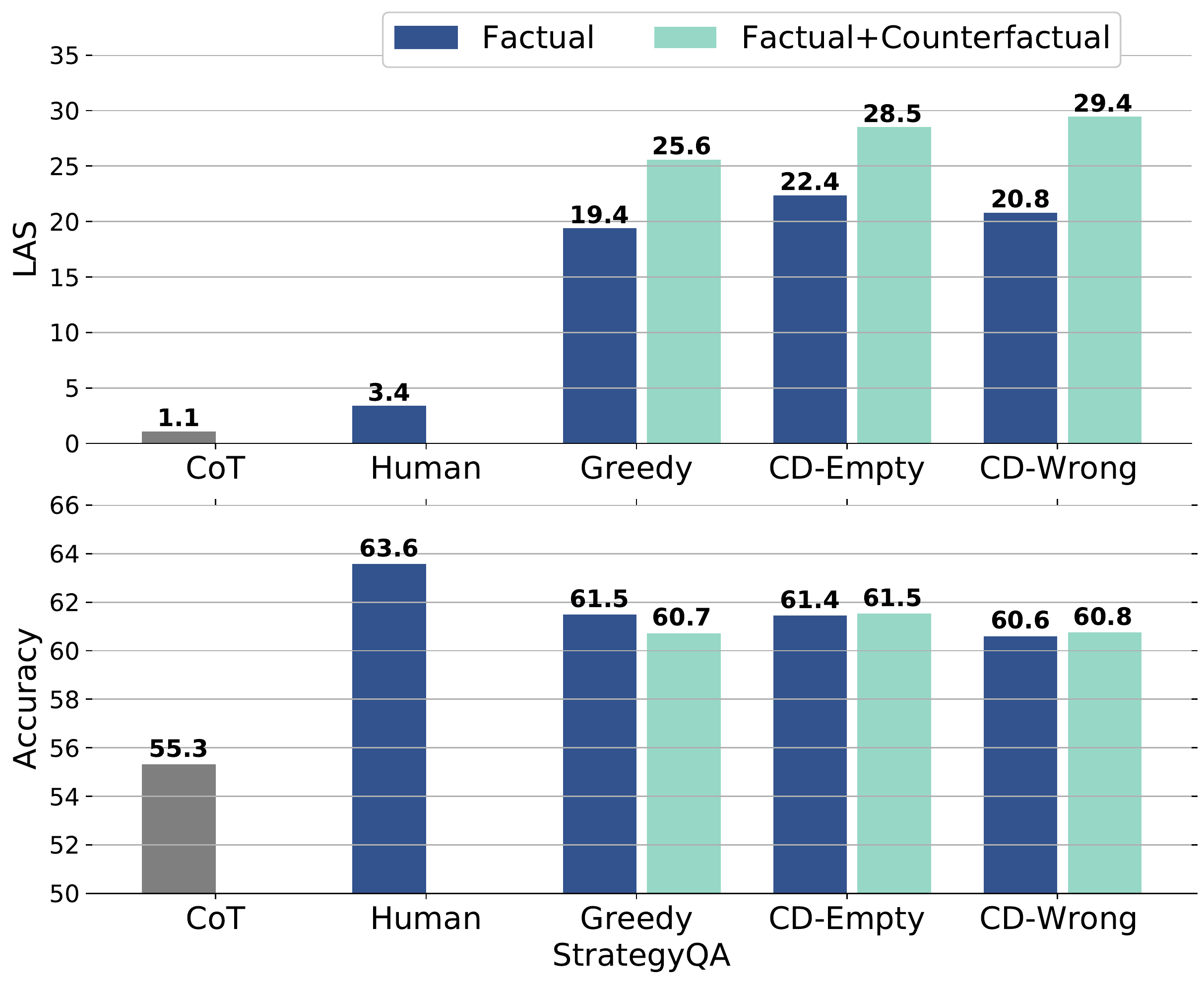}  
  \label{fig:acc_las_strategyqa}
\end{subfigure}
\begin{subfigure}{.45\textwidth}
  \centering
  \includegraphics[width=\linewidth]{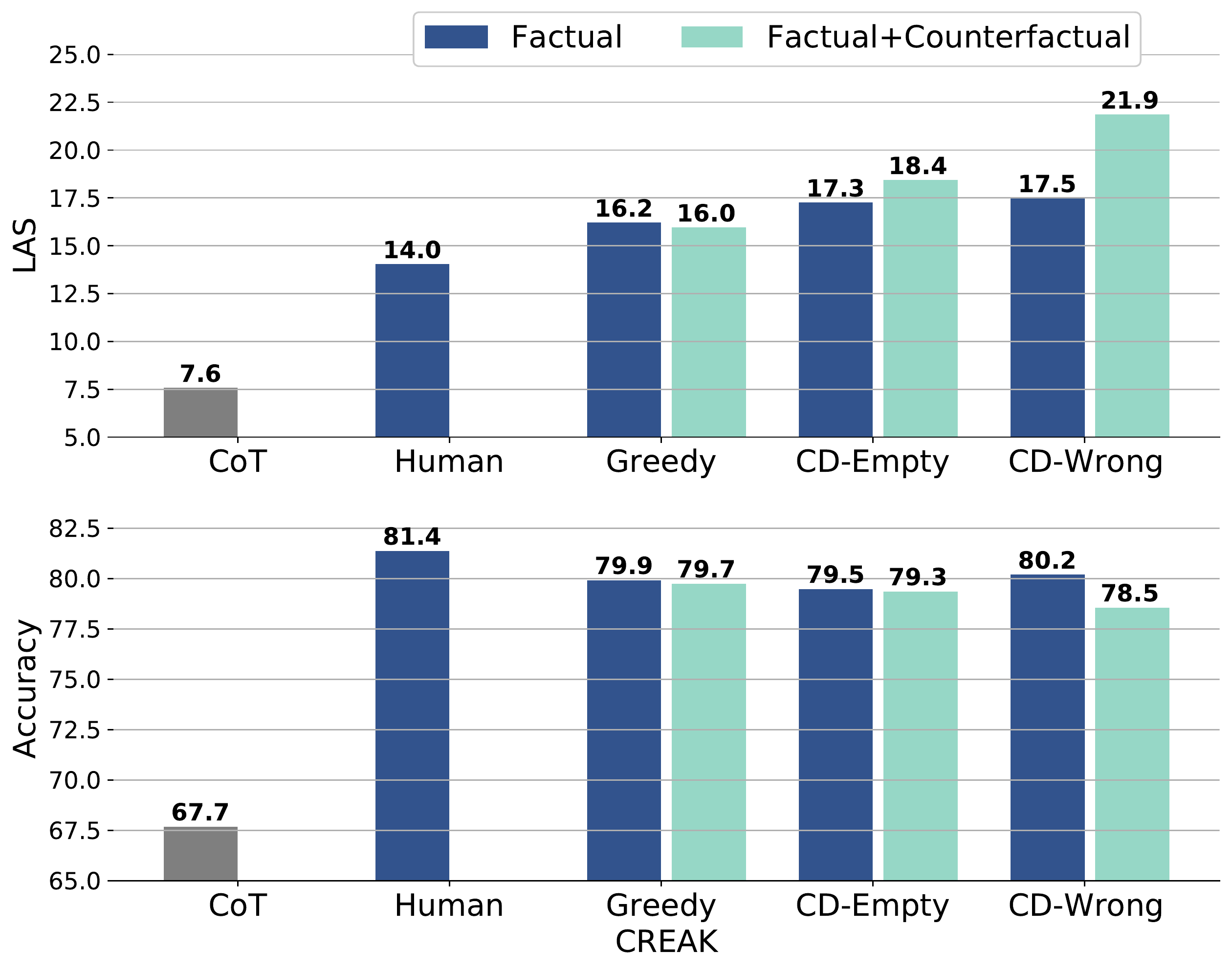}  
  \label{fig:acc_las_creak}
\end{subfigure}
\begin{subfigure}{.45\textwidth}
  \centering
  \includegraphics[width=\linewidth]{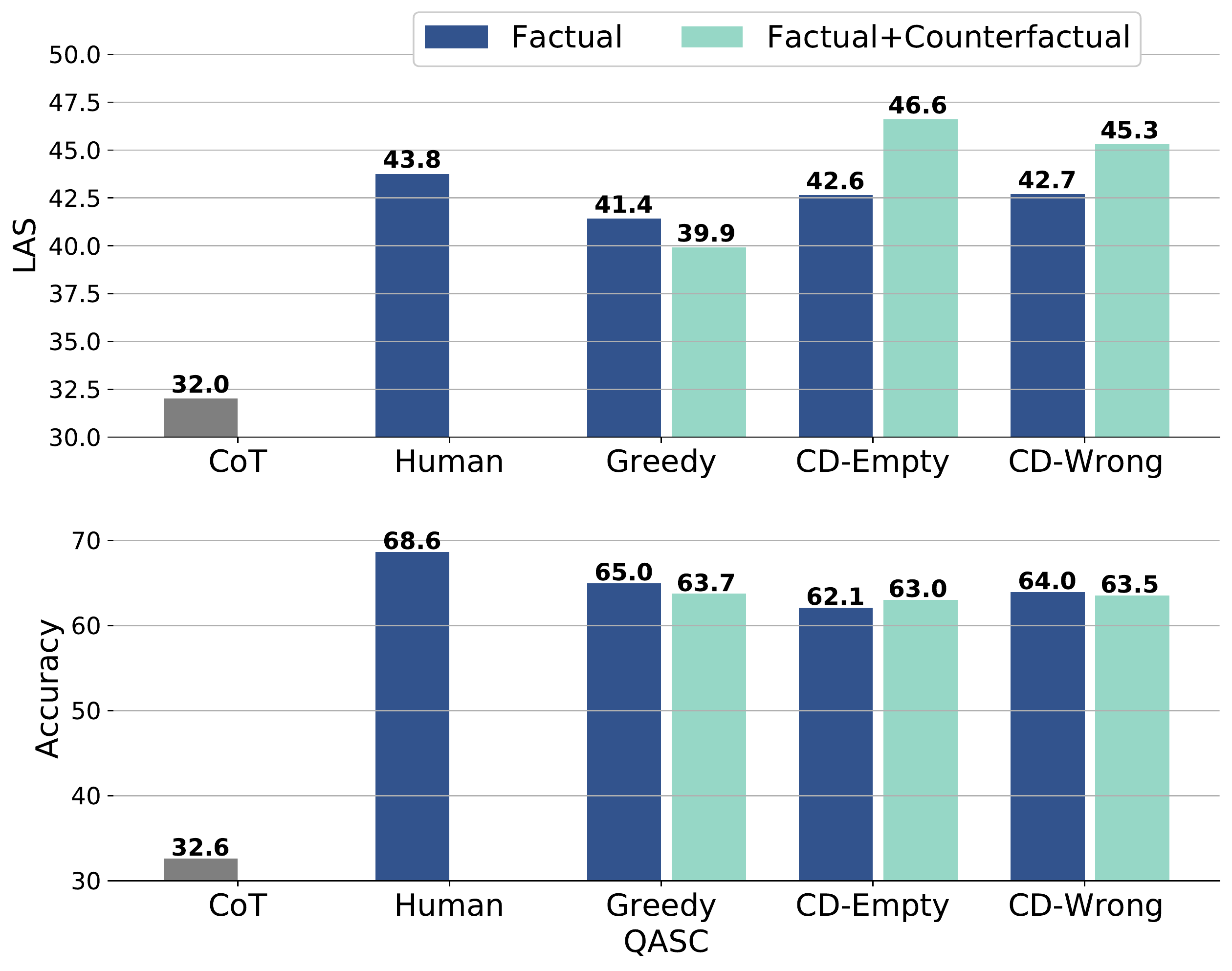}  
  \label{fig:acc_las_qasc}
\end{subfigure}
    \caption{Faithfulness (LAS) and task performance (accuracy) of the compared methods on the experimented datasets. The x-axis represents the CoT baseline and knowledge distillation methods that use Human Annotation, Greedy Decoding, and Contrastive Decoding with empty strings/wrong answers as the teachers. For knowledge distllation methods, we apply Factual training and Factual+Counterfactual training to train the students.}
    \label{fig:acc_las}
\end{figure*}

\smallskip
\noindent
\textbf{Can contrastive decoding lead to a more consistent teacher?}
~Figure~\ref{fig:teacher_las} shows the consistency between the rationales generated by different teachers and the gold answers measured by LAS. Across four datasets, contrastive decoding with either empty or wrong answers yield more consistent rationales compared to human annotation and greedy decoding. This demonstrates the effectiveness of our contrastive decoding strategy in encouraging the teacher to generate more on-topic rationales. Moreover, using wrong answers is better than using empty strings for contrastive decoding. This shows that by contrasting with the wrong answers, the teacher can generate more distinguishable rationales that lead to the gold answers, thus obtain higher consistency. Greedy decoding yields less consistent rationales compared to human annotation, verifying our claim that LMs are prone to generating text not grounded by the gold answers.
\begin{table}[t]
\centering
\caption{Human evaluation on the rationales generated by different teacher models for StrategyQA. A fair level of agreement measured by Fleiss Kappa ($\kappa$=0.26) is obtained among three annotators.}
\label{tab:human_eval}
\scalebox{0.7}{
\begin{tabular}{l|ccc}
\toprule
Teacher Model & Grammaticality &  New Info & Supports Answer \\
\midrule
Greedy & \textbf{0.99} & 0.65 & 0.48 \\
Contrast.-Empty & 0.97 & 0.77 & 0.58 \\
Contrast.-Wrong & 0.97 & \textbf{0.82} & \textbf{0.63}\\
\bottomrule
\end{tabular}
}

\end{table}

We also conduct a human evaluation over 100 rationales generated by different decoding strategies for StrategyQA. Annotators are asked to judge the rationales by 3 dimensions: 1) Grammaticality (Is the rationale grammatical?) 2) New Info (Does the rationale provide new
information not expressed in the question?) 3) Supports Answer (Does
the rationale justify the answer?). Table~\ref{tab:human_eval} confirms that our two contrastive decoding strategies yield more informative and on-topic rationales than greedy decoding, with a slightly worse grammaticality. We list examples in Table~\ref{tab:case_study} (appendix) to showcase how rationales from contrative decoding are more consistent with gold answers than greedy decoding.

\smallskip
\noindent
\textbf{Can a more consistent teacher train a more faithful student?}
Figure~\ref{fig:acc_las} (upper parts of each sub-figure) shows the faithfulness of the students measured in LAS on the experimented datasets. First, the CoT method often achieves much lower LAS compared to the KD methods across four datasets, showing that the generated rationales do not faithfully reflect the decision making in CoT. Second, we observe that students trained with the rationales from contrastive decoding with either empty strings or wrong answers generally achieve higher LAS scores compared to the baselines. Together with the observation on the consistency of the teacher (Figure~\ref{fig:teacher_las}), this validates that a more consistent teacher train a more faithful student and the inconsistency in the training data generated by the teacher will be inherited by the student. 

\smallskip
\noindent
\textbf{Can couterfactual reasoning loss further improve the faithfulness?} Figure~\ref{fig:acc_las} shows the students fine-tuned additionally with counterfactual training loss achieve higher faithfulness than their counterparts which are fine-tuned with factual training only. This validates that counterfactual reasoning can further improve the student's faithfulness, as it may still treat rationale generation and answer prediction as two independent processes. 

\smallskip
\noindent
\textbf{Can a faithful student still preserve its performance?}~Figure~\ref{fig:acc_las} (lower parts of each sub-figure) shows the performance of the students measured in accuracy. First, CoT methods achieve lower accuracy compared to the KD methods, showing the benefit of combining the supervision from the teacher (the rationales) and the labeled datasets (the answers). Second, all the KD methods achieve comparable performance. Together with the observation over faithfulness, this demonstrates our method can improve faithfulness of the model while not hurting its performance. Note that the student which learns from human annotation achieves slightly better results compared to other students. This is because the human rationales are less consistent with the answers (as evidenced in Figure~\ref{fig:teacher_las}). Therefore, the student learns to generate the rationales and predict the answers more independently, which allows it to exploit the spurious correlation and achieve better performance. Our further analysis (\S~\ref{sec:rationale_perturb}) shows that such performance gain is suspicious as changing the rationales does not change the student's predictions mostly.

\subsection{Ablation on the student model size}
\begin{figure}[t]
    \centering
    \includegraphics[width=0.45\textwidth]{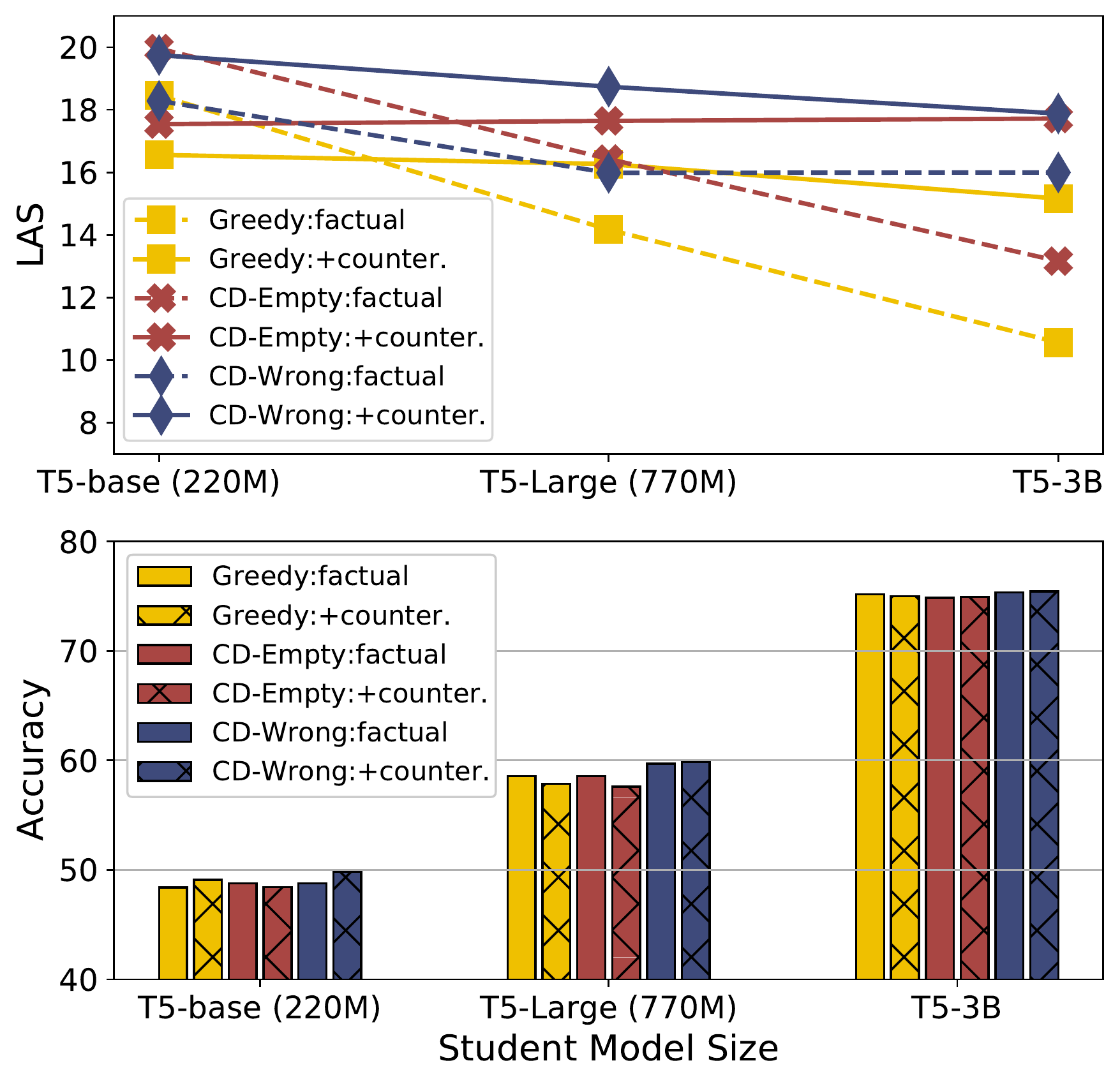}
\vspace{-0.1cm}
    \caption{Faithfulness (LAS) and task performance (accuracy) of the compared methods with different student model sizes. Each model is named by the teacher it learns from and the training objective as \texttt{teacher model:training objective}.
    }
    \label{fig:ablation_student_size}
\vspace{-0.5cm}
\end{figure}

We ablate the student model size to see how its faithfulness and performance are affected. From Figure~\ref{fig:ablation_student_size}, we observe that larger student models achieve higher performance but lower faithfulness. This confirms that it requires sufficient capacity for storing knowledge necessary for reasoning~\cite{wei2022emergent}, but larger models are also better at answering the questions independently of the rationales. Still, our models are more faithful than baselines and comparable in performance with different model sizes.

\begin{figure*}[t]
    \centering
\begin{subfigure}{.46\textwidth}
  \centering
  \includegraphics[width=\linewidth]{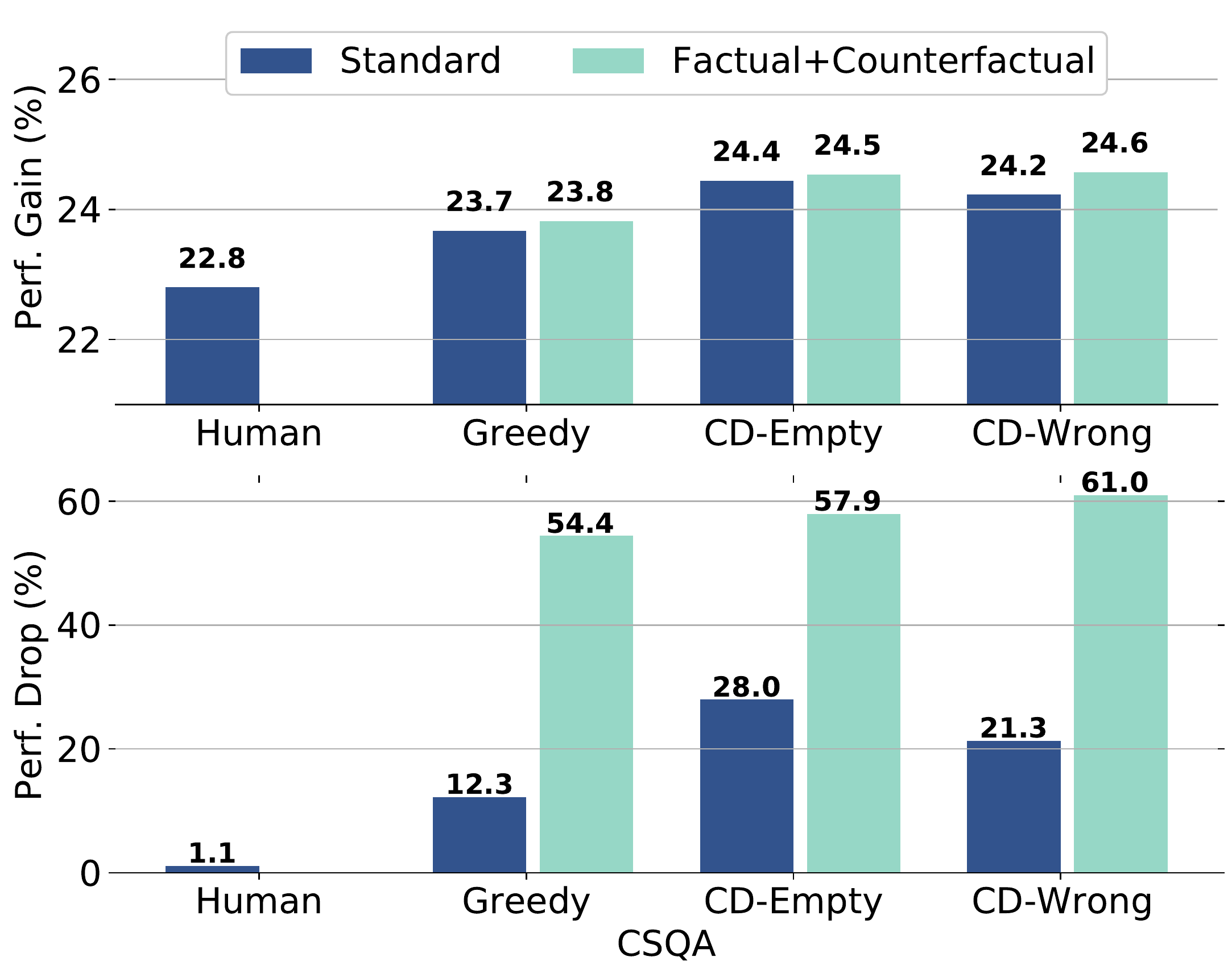}  
  \label{fig:perf_change_csqa}
\end{subfigure}
\begin{subfigure}{.46\textwidth}
  \centering
  \includegraphics[width=\linewidth]{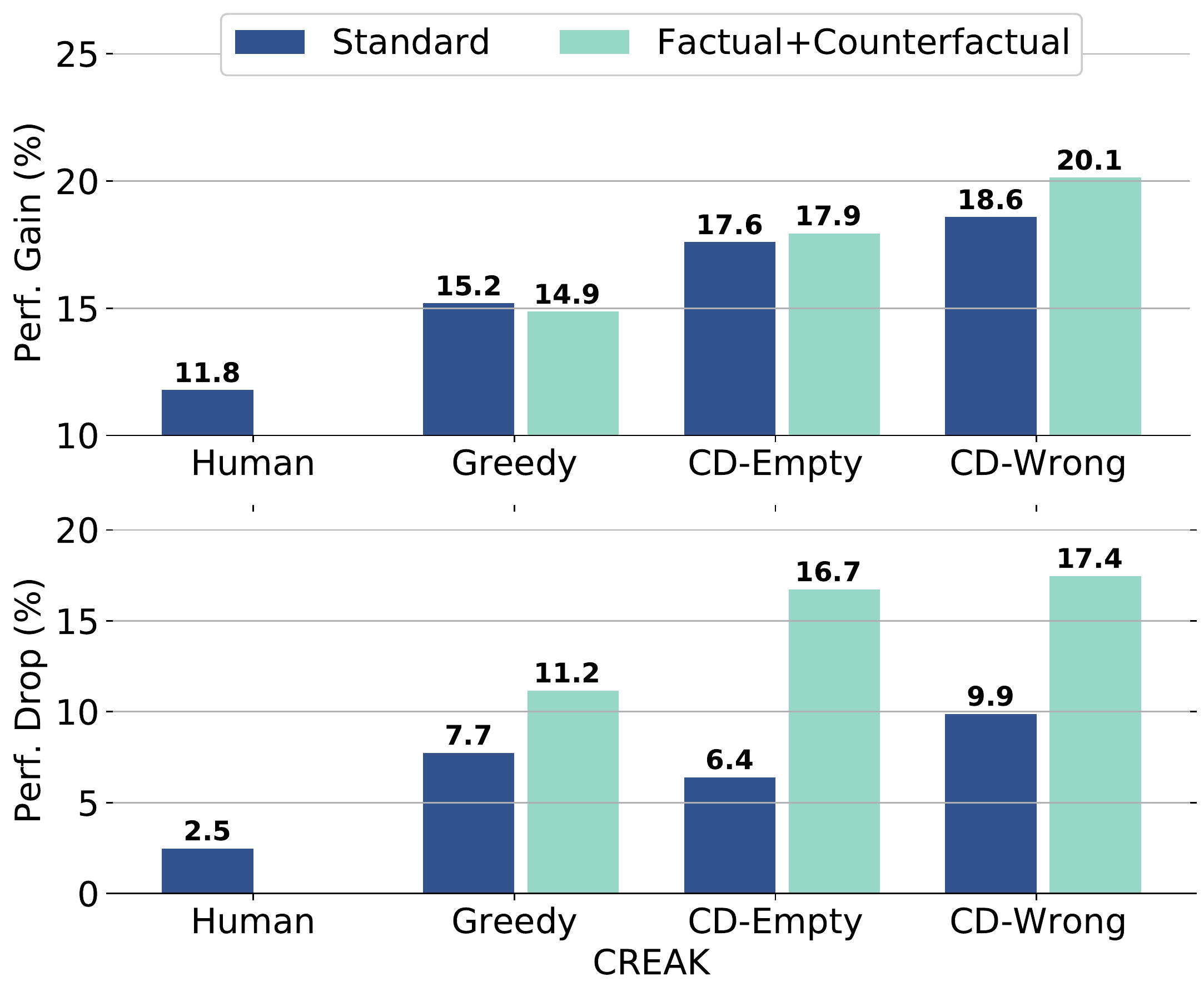}  
  \label{fig:perf_change_creak}
\end{subfigure}
    \caption{Performance gain (drop) of the compared methods when the oracle (perturbed) rationales are fed to the decoder of the model on CSQA and CREAK.}
    \label{fig:perf_change}
\end{figure*}

\subsection{Controlling the behavior of the Student}\label{sec:rationale_perturb}
One important utility of faithful rationales is that we can have more control over the behavior of the student via changing its rationales. If the model can make predictions consistent with its rationales, we can either impair or improve the its performance by perturbing or refining its rationales. To verify this, we conduct two types of edition to the rationales generated by the student, namely perturbation and refinement as described below. We then feed the edited rationales to the decoder of the student directly (as teacher forcing) and see if the student will act accordingly, i.e., predict more badly (or accurately) due to the worse (or better) rationales.

\smallskip
\noindent
\textbf{Rationales Perturbation}
For perturbing the rationales, we randomly replace $50\%$ of the tokens in the generated rationales from the student and then feed the perturbed rationales $r{'}$ back to the decoder of the student. We finally calculate the performance drop (or sensitivity), i.e., $Acc(qr\rightarrow a^*)-Acc(qr^{'}\rightarrow a^*)$. Figure~\ref{fig:perf_change} (the lower parts) shows the results on CSQA and CREAK. First, perturbing the rationales from the student that is fine-tuned with human-annotation has little (down to $1.1\%$ on CSQA) impact on its performance, meaning that the student largely ignores the rationales when making prediction. Second, learning from rationales obtained by contrastive decoding with empty or wrong answers leads to a student that is more sensitive to the rationale perturbation compared to learning from greedy decoding. This again verifies the necessity of having a consistent teacher in order to train a faithful student. Lastly, our counterfactual training loss further improves the sensitivity of the student, demonstrating that the student is more faithful towards the rationales.

\smallskip
\noindent
\textbf{Rationales Refinement}
As a proxy refinement, we obtain the oracle rationales $r^{*}$ automatically by asking the teacher to rationalize for gold answers using each compared decoding strategy. For the student trained with human annotation, we directly use the annotated rationales as the oracle. We then calculate the performance gain, i.e., $Acc(qr^{*}\rightarrow a^*)-Acc(qr\rightarrow a^*)$. Figure~\ref{fig:perf_change} (the upper parts) shows the results on CSQA and CREAK. First, we observe that oracle human-annotated rationales do not bring as much performance gain as machine-generated rationales do. This demonstrates that even trained with human annotation, the student is still prone to being unfaithful to its rationales. 
Second, we observe that contrastive decoding (with either empty strings or wrong answers) leads to higher performance gains from the student. By adding counterfactual training, the performance gains are further increased. This demonstrates the advantage brought by our method, which is that we can have more success in debugging a reasoning model by refining its rationales.

\section{Related Works}

\smallskip
\noindent
\textbf{Free-text Rationales}~
A variety of datasets have been proposed to collect human-annotated rationales alongside each task instance~\cite{camburu2018snli,rajani-etal-2019-explain,aggarwal2021explanations}, aiming to train the downstream models to explain their predictions in natural language. However, human annotation is expensive and the resulting rationales are reported to be of poor quality~\cite{aggarwal2021explanations,sun2022investigating}. Our work leverages a prompted LM to obtain rationales automatically for supporting both correct and incorrect answers, using only a few annotated examples as demonstration. The rationales for supporting the incorrect answers further enable the student to conduct counterfactual reasoning, which is not available from existing human annotation.

\smallskip
\noindent
\textbf{Prompted Self-Rationalization Models}~Recent works have been proposed to prompt large LMs to generate a free-text rationale before making the prediction~\cite{nye2021show,wei2022chain}. However, this technique relies on extremely large LMs (with over 100B parameters) to work effectively~\cite{wei2022chain,wei2022emergent}, which requires significant computation resources or expensive API calls~\cite{shridhar2022distilling}. Meanwhile, the rationales generated by such models are shown to contradict the context~\cite{ye2022unreliability} and fail to faithfully represent the underlying reasoning process~\cite{wang2022pinto}. In contrast, our student is trained to be more faithful towards its generated rationales using a smaller LM.

\smallskip
\noindent
\textbf{Knowledge Distillation}
There exist some works that explore the idea of distilling rationales knowledge from a large LM to a small LM as the student. 
\citeauthor{chan2022knife} proposed to learn a student model that only predicts answers from a teacher model that is augmented with rationales.
\citeauthor{eisenstein2022honest} proposed to train the student to extract the sentence containing the answer, which is not applicable to reasoning tasks that require background knowledge. 
\citeauthor{shridhar2022distilling} proposed to train the student to ask and answer sub-questions necessary for decomposing the main question, which is tailored to solve math word problems~\cite{cobbe2021training} with an equation generator for guiding the student while we do not have such a constraint. \citeauthor{li2022explanations} proposed to train the student on the joint task of generating the answers and the rationales, which only act as a regularization and do not affect the student's prediction during inference. More importantly, both \citeauthor{shridhar2022distilling} and \citeauthor{li2022explanations} do not consider the faithfulness of the rationales, which is critical for examining the behavior of the student.
\section{Conclusion}
This work presents a faithful KD framework for  learning a small, self-consistent CoT model from a large teacher model. To ensure the student reason faithfully, we propose (1) contrastive decoding for obtaining a consistent teacher and (2) counterfactual reasoning for teaching a faithful student. Experiments show that these two techniques jointly lead to a more faithful student compared to the baselines, while preserving much performance accuracy. Our further analysis shows that changing the rationales has a larger impact on the student's behavior and thus we can have more success in debugging the model by refining its rationales.
\section*{Limitations}
Compared to a standard knowledge distillation process, our method requires additional computation when preparing training data and training the student. First, our contrastive decoding needs to perform forward pass in the teacher model one time more than greedy decoding does to obtain the perturbed plausibility for each token generated (Eq.~\ref{eq:contrastive_decoding}). Second, our KD process introduces additional training data for training the student with the counterfactual reasoning objective (Eq.\ref{eq:counterfactual_training}). Besides computation cost, this work focuses on improving faithfulness of the rationales rather than performance, which is complementary to prior works which leverages rationales for improving the performance only. 

\section*{Ethics Statement}
Our KD process leverages large LMs to obtain rationale annotation, which may expose social bias encoded in these models~\cite{lucy-bamman-2021-gender}. The bias may be further inherited by the student model. Nevertheless, our method improves the faithfulness of the rationales, making the predictions from the student accountable. Without the faithful rationales, it would be unclear to users about whether the model is making predictions based on some unintended bias.
\bibliography{custom}
\bibliographystyle{acl_natbib}

\appendix
\section{Appendix}
\subsection{Example generations from teachers}
\begin{table}[H]
\centering
\caption{Examples where rationales generated by contrastive decoding with empty/wrong answers are rated higher than rationales generated by greedy decoding in our human evaluation.
}
\label{tab:case_study}
\small
\begin{tabular}{p{7.8cm}}
\toprule
\textbf{Q}: \color{blue}Do workers at Nissan's headquarters eat with chopsticks?\\
\textbf{A}: \color{green} Yes. \\
\textbf{Greedy}: Workers at Nissan's headquarters eat with chopsticks. Chopsticks are used in China. Thus, workers at Nissan's headquarters eat with chopsticks. \\
\textbf{Contrastive-Empty}: Workers at Nissan's headquarters are Japanese. Japanese eat with chopsticks. \\
\textbf{Contrastive-Wrong}: Nissan's headquarters is in Japan. Japan is a country where people eat with chopsticks. Thus, workers at Nissan's headquarters eat with chopsticks.\\
\midrule
\textbf{Q}: \color{blue}Would Constitution of the United States paper offend PETA? \\
\textbf{A}: \color{green} Yes.\\
\textbf{Greedy}: PETA is an animal rights group. The Constitution of the United States is a document that protects the rights of animals. Thus, PETA would be offended by the Constitution of the United States.\\
\textbf{Contrastive-Empty}: PETA is against the use of animal products. The Constitution of the United States is written on animal products. Thus, PETA would be offended by the Constitution of the United States. \\
\textbf{Contrastive-Wrong}: PETA is against the use of animal products. The Constitution of the United States is written on animal products. Thus, PETA would be offended by the Constitution of the United States.\\
\midrule
\textbf{Q}: \color{blue}Does Sam Harris worship Shiva? \\
\textbf{A}: \color{green} No. \\
\textbf{Greedy}:  Sam Harris is a Hindu. Shiva is a Hindu god. Thus, Sam Harris does not worship Shiva. \\
\textbf{Contrastive-Empty}: Shiva is a Hindu god. Sam Harris is an atheist. Thus, Sam Harris does not worship Shiva. \\
\textbf{Contrastive-Wrong}: Shiva is a Hindu god. Sam Harris is an atheist. Thus, Sam Harris does not worship Shiva. \\
\midrule
\textbf{Q}:  \color{blue} Are a dozen pickles good for easing hypertension?\\
\textbf{A}: \color{green} No. \\
\textbf{Greedy}: A dozen pickles have about 1,000 calories. A person needs about 2,000 calories a day. Thus, a dozen pickles would not be good for easing hypertension.\\
\textbf{Contrastive-Empty}: Pickles are acidic. Acidic foods are not good for easing hypertension. \\
\textbf{Contrastive-Wrong}: The recommended daily intake of sodium is 2,400 mg. A dozen pickles have about 0.5 g of sodium. Thus, a dozen pickles would not be good for easing hypertension.\\
\midrule
\textbf{Q}: \color{blue} Would it be unusual to see frost in September in Texas?\\
\textbf{A}: \color{green} Yes.\\
\textbf{Greedy}: Frost can occur in September in Texas. However, it is unusual.\\
\textbf{Contrastive-Empty}:  Frost occurs in the winter. Thus, it would be unusual to see frost in September in Texas.\\
\textbf{Contrastive-Wrong}: Frost is usually seen in the winter. Thus, it would be unusual to see frost in September in Texas.\\
\bottomrule
\end{tabular}

\end{table}

\end{document}